\documentclass[runningheads]{llncs}
\usepackage{graphicx}
\usepackage{amsmath}
\usepackage{amssymb}
\usepackage{marvosym}
\usepackage{bm}
\usepackage{subfigure}
\usepackage{multirow}
\usepackage{makecell}
\usepackage{latexsym}
\usepackage{hyperref}
\hypersetup{
    colorlinks = true,
    citecolor = blue
}

\begin{document}

\title{Partial Adversarial Domain Adaptation}

\titlerunning{Partial Adversarial Domain Adaptation}

\authorrunning{Zhangjie Cao, Lijia Ma, Mingsheng Long, and Jianmin Wang}

\author{Zhangjie Cao 
\and Lijia Ma
\and Mingsheng Long(\Letter)
\and Jianmin Wang
}
\authorrunning{Zhangjie Cao, Lijia Ma, Mingsheng Long, and Jianmin Wang}

\institute{School of Software, Tsinghua University, China\\
National Engineering Laboratory for Big Data Software\\
Beijing National Research Center for Information Science and Technology\\
\email{\{caozhangjie14,malijia15\}@gmail.com, \{mingsheng,jimwang\}@tsinghua.edu.cn}}

\maketitle

\begin{abstract}
Domain adversarial learning aligns the feature distributions across the source and target domains in a two-player minimax game. Existing domain adversarial networks generally assume identical label space across different domains. In the presence of big data, there is strong motivation of transferring deep models from existing big domains to unknown small domains. This paper introduces partial domain adaptation as a new domain adaptation scenario, which relaxes the fully shared label space assumption to that the source label space subsumes the target label space. Previous methods typically match the whole source domain to the target domain, which are vulnerable to negative transfer for the partial domain adaptation problem due to the large mismatch between label spaces. We present Partial Adversarial Domain Adaptation (PADA), which simultaneously alleviates negative transfer by down-weighing the data of outlier source classes for training both source classifier and domain adversary, and promotes positive transfer by matching the feature distributions in the shared label space. Experiments show that PADA exceeds state-of-the-art results for partial domain adaptation tasks on several datasets.
\end{abstract}

\section{Introduction}
Deep neural networks have made significant advances to a variety of machine learning problems and applications. However, the significant advances attribute to the availability of large-scale labeled data. Since manually labeling sufficient training data for various applications is often prohibitive, for problems short of labeled data, there is strong incentive to designing versatile algorithms to reduce the labeling consumption. Domain adaptation methods~\cite{cite:TKDE10TLSurvey} enable the ability to leverage labeled data from a different but related source domain. At the core of these methods is the shift in data distributions across different domains, which hinders the generalization of predictive models to new target tasks~\cite{cite:COLT09DAT}.

Existing domain adaptation methods generally assume that the source and the target domains share identical label space but follow different distributions. These methods close the large gap between different domains by learning domain-invariant feature representations using both domain data but without using target labels, and apply the classifier trained on the source domain to the target domain. Recent research has shown that deep networks can disentangle explanatory factors of variations underlying domains to learn more transferable features for domain adaptation~\cite{cite:CVPR13MidLevel,cite:ICML14DeCAF,cite:NIPS14CNN}. Along this line, domain adaptation modules such as moment matching \cite{cite:Arxiv14DDC,cite:ICML15DAN,cite:NIPS16RTN,cite:ICML17JAN} and adversarial adaptation~\cite{cite:ICML15RevGrad,cite:ICCV15SDT,cite:CVPR17ADDA} have been embedded in deep networks to learn domain-transferable representations.

\begin{figure}[tbp]
  \centering
  \includegraphics[width=0.9\textwidth]{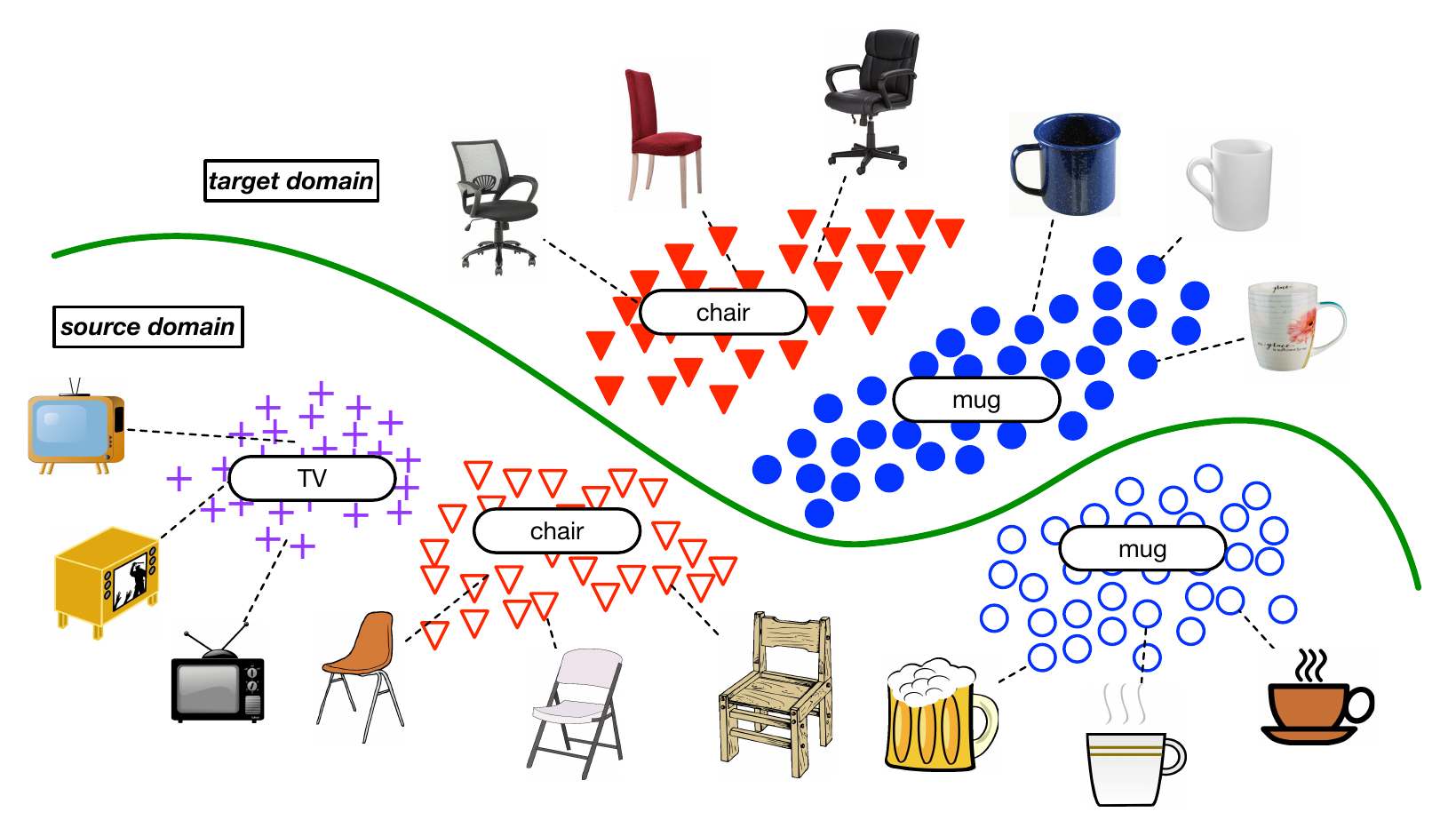}
  \caption{The partial domain adaptation scenario introduced in this paper, where source label space (`TV', `chair', `mug') is a superset of target label space (`chair', `mug'). This scenario is difficult as the source domain may have some outlier classes not appearing in the target domain, e.g. the `TV' class. These outlier source classes will make the well-known negative transfer bottleneck more prominent. Another technical difficulty is that it is nontrivial to identify which classes are outlier source classes since the target classes are unknown during training. In this paper, we will tackle these challenges in an end-to-end deep learning framework, Partial Adversarial Domain Adaptation (PADA).}
   \label{fig:PADAproblem}
\end{figure}

However, in real applications, it is usually not easy to find a source domain with identical label space as the target domain of interest. Thus, previous domain adaptation methods can hardly fit into proper datasets to train a domain-invariant model. Also, it is really cumbersome to seek for new source domains for emerging target domains. 
Thanks to the big data evolution, large-scale datasets with rich supervised information such as ImageNet-1K become accessible. As a common practice, ImageNet-1K is used as a universal repository to train deep models that are later fine-tuned to a variety of significantly different tasks. However, until now we can only reuse the learned features.
And a natural ambition is to further transfer the classification layers of deep networks from a large dataset with supervision e.g. ImageNet, to a small target dataset without supervision e.g. Caltech-256. Since the large dataset is required to be big enough, it is reasonable to assume that its label space subsumes that of our target dataset.

Towards the aforementioned ambition, we introduce a novel \emph{partial domain adaptation} problem, which assumes that the target label space is a subspace of the source label space.  As shown in Fig.~\ref{fig:PADAproblem}, this novel scenario is more general and challenging than standard domain adaptation, since the \emph{outlier} source classes (`TV') will trigger negative transfer when discriminating the target classes (`chairs' and `mug'). Negative transfer is the worst case that an adapted learner performs even worse than a supervised classifier trained solely on the source domain, which is the key bottleneck of domain adaptation to be widely adopted by practical applications \cite{cite:TKDE10TLSurvey}.
Thus, matching the whole source and target domains as previous methods is not an effective solution to this new problem.

In this paper, we present Partial Adversarial Domain Adaptation (PADA), an end-to-end framework that largely extends the ability of domain adversarial adaptation approaches \cite{cite:ICML15RevGrad,cite:ICCV15SDT,cite:CVPR17ADDA} to address the new partial domain adaptation scenario. PADA aligns the feature distributions of the source and target data in the shared label space and more importantly, identifies the irrelevant source data belonging to the outlier source classes and down-weighs their importance automatically. The key improvement over previous methods is the capability to simultaneously promote positive transfer of relevant source data and alleviate negative transfer of irrelevant source data. Experiments show that our models exceed state-of-the-art results for partial domain adaptation on public datasets.

\section{Related Work}

Domain adaptation \cite{cite:TKDE10TLSurvey} bridges different domains following different distributions to mitigate the burden of manual labeling for machine learning~\cite{cite:TNN11TCA,cite:TPAMI12DTMKL,cite:ICML13TCS,cite:NIPS14FTL}, computer vision \cite{cite:ECCV10Office,cite:CVPR12GFK,cite:NIPS14LSDA} and natural language processing \cite{cite:JMLR11MTLNLP}. Supervised domain adaptation~\cite{cite:ICCV15SDT,cite:CVPR17DAMA,cite:ICCV17SDA,cite:NIPS17FADA} exploits a few labeled data in the target domain. While supervised domain adaptation achieves significant performance, unsupervised domain adaptation~\cite{cite:Arxiv14DDC,cite:ICML15DAN,cite:CVPR17ADDA,cite:ICML15RevGrad,cite:NIPS16RTN,cite:ICML17JAN} is more practical since no labeled data is required. We focus on unsupervised domain adaptation in this paper. 

Deep networks disentangle different explanatory factors of variations in the learned representations \cite{cite:TPAMI13DLSurvey} and manifest invariant factors underlying different populations that transfer well across similar tasks \cite{cite:NIPS14CNN}. Therefore, we mainly focus on deep domain adaptation methods, which enables domain adaptation by reducing the distribution discrepancy of deep features across different domains and have been proved to yield state-of-the-art performance on several domain adaptation tasks. Maximum Mean Discrepancy (MMD) based methods~\cite{cite:Arxiv14DDC,cite:ICML15DAN,cite:ICML17JAN} transfers deep convolutional networks (CNNs) by adding adaptation layers through which the kernel embeddings of distributions are matched by minimizing MMD. Residual transfer network~\cite{cite:NIPS16RTN} improves the MMD-based methods by adding a shortcut path and adopting entropy minimization criterion.

Driven by the popularity of generative adversarial networks (GANs), several methods~\cite{cite:ICML15RevGrad,cite:CVPR17ADDA} add a subnetwork as a domain discriminator on the last feature layers to discriminate features of different domains, while the deep features are learned to deceive the domain discriminator in a two-player game. Label Efficient Learning~\cite{cite:NIPS17LEL} addresses different label spaces by extending the entropy minimization criterion~\cite{cite:NIPS16RTN} over the pairwise similarity of a target image with each source image, enforcing each target image to be similar to only a few source images.  

These methods may be restricted by the assumption that the source and target domains share the same label space, which does not hold in partial domain adaptation. Adaptive Deep Learning~\cite{cite:ICRA18LoAd}, somehow reduces the negative transfer of outlier classes by localizing the image regions more responsible for the domain shift as well as the regions that are shared among domains to guide the attention of the classifier. But for images with no regions related to the target domain, the attention mechanism may fail by wrongly localizing related regions. 

\section{Partial Adversarial Domain Adaptation}
This paper introduces \emph{partial domain adaptation}, a novel domain adaptation scenario where the source domain label space $\mathcal{C}_t$ is a superset of the target domain label space $\mathcal{C}_s$ i.e. $\mathcal{C}_t \subset \mathcal{C}_s$. This scenario generalizes standard domain adaptation with identical label spaces, and can be widely applied to real applications, since with the availability of big data, it is not difficult to find a large-scale dataset (e.g. ImageNet) and adapt our model trained on that dataset to any small-scale dataset of interest (e.g. Caltech-256), given the partial assumption holds. By this means, we can avoid burdensome work to provide supervised information for the target dataset.

Similar to standard domain adaptation, in partial domain adaptation we are also provided with a \emph{source} domain $\mathcal{D}_s = \{(\mathbf{x}_i^s,y^s_i)\}_{i=1}^{n_s}$ of $n_s$ labeled examples associated with $|\mathcal{C}_s|$ classes and a \emph{target} domain ${{\mathcal D}_t} = \{ {\mathbf{x}}_i^t\} _{i = 1}^{{n_t}}$ of $n_t$ unlabeled examples associated with $|\mathcal{C}_t|$ classes, but differently, we have $|\mathcal{C}_s| > |\mathcal{C}_t|$ in partial domain adaptation. The source and target domains are sampled from distributions $p$ and $q$ respectively. While in standard domain adaptation we have $p \ne q$, in partial domain adaptation, we further have $p_{\mathcal{C}_t} \ne q$, where $p_{\mathcal{C}_t}$ denotes the distribution of the source domain labeled data belonging to label space $\mathcal{C}_t$.
The goal of this paper is to design a deep neural network that enables learning of transferable features $\mathbf{f} = G_f\left( {\bf{x}} \right)$ and adaptive classifier $y = G_y\left( {\bf{f}} \right)$ to close the domain gap, such that the target risk ${\Pr _{\left( {{\mathbf{x}},y} \right) \sim q}}\left[ {G_y \left( G_f({\mathbf{x}}) \right) \ne y} \right]$ can be bounded by minimizing the source domain risk and the cross-domain discrepancy.

In standard domain adaptation, one of the main difficulties is that the target domain has no labeled data and thus the source classifier $G_y$ trained on source domain $\mathcal{D}_s$ cannot be directly applied to target domain $\mathcal{D}_t$, due to the distribution discrepancy of $p \ne q$. In partial domain adaptation, another more difficult challenge is that we even do not know which part of the source domain label space $\mathcal{C}_s$ is shared with the target domain label space $\mathcal{C}_t$, because $\mathcal{C}_t$ is not known during training. This results in two technical difficulties.
On one hand, the source domain labeled data belonging to \emph{outlier} label space $\mathcal{C}_s \backslash \mathcal{C}_t$ will cause \textbf{negative transfer} effect to the overall performance. 
Existing deep domain adaptation methods \cite{cite:ICML15DAN,cite:ICML15RevGrad,cite:ICCV15SDT,cite:NIPS16RTN} generally assume source domain and target domain have the same label space and match the whole distributions $p$ and $q$, which are prone to negative transfer since the source and target label spaces are different and thus the outlier classes should not be matched.
Thus, how to undo or at least decay the influence of the source labeled data in outlier label space $\mathcal{C}_s \backslash \mathcal{C}_t$ is the key to mitigating negative transfer.
On the other hand, reducing the distribution discrepancy between $p_{\mathcal{C}_t}$ and $q$ is crucial to enabling \textbf{positive transfer} in the shared label space $\mathcal{C}_t$.
These two interleaving challenges should be tackled jointly through filtering out the negative influence of unrelated part of source domain and meanwhile enabling effective domain adaptation between related part of source domain and target domain.

In summary, we should tackle two challenges to enabling partial domain adaptation. \textbf{(1)} Mitigate negative transfer by filtering out unrelated source labeled data belonging to the outlier label space $\mathcal{C}_s \backslash \mathcal{C}_t$. \textbf{(2)} Promote positive transfer by maximally matching the data distributions $p_{\mathcal{C}_t}$ and $q$ in the shared label space $\mathcal{C}_t$. We propose a partial domain adversarial network to address both challenges.

\subsection{Domain Adversarial Neural Network}
Domain adaptation is usually reduced to matching the feature distributions of the source and target domains. In the deep learning regime, this can be done by learning new feature representations such that the source and target domains are not distinguishable by a domain discriminator. This idea leads to the a series of domain adversarial neural networks (DANN)~\cite{cite:ICML15RevGrad,cite:ICCV15SDT}, achieving strong performance in standard domain adaptation with shared label space across domains. More formally, DANN is a two-player minimax game, where the first player is a domain discriminator $G_d$ trained to distinguish the source domain from the target domain, and the second player is a feature extractor $G_f$ simultaneously trained to confuse the domain discriminator.

In order to extract domain-transferable features $\mathbf{f}$, the parameters $\theta_f$ of the feature extractor $G_f$ are learned by maximizing the loss of domain discriminator $G_d$, while the parameters $\theta_d$ of domain discriminator $G_d$ are learned by minimizing the loss of the domain discriminator. In addition, the loss of source classifier $G_y$ is also minimized to guarantee lower source domain classification error. The overall objective of the Domain Adversarial Neural Network (DANN) \cite{cite:ICML15RevGrad} is
\begin{equation}\label{eqn:GRL}
\begin{aligned}
	C_{0} \left( {{\theta _f},{\theta _y},{\theta _d}} \right) & = \frac{1}{{{n_s}}}\sum\limits_{{{\mathbf{x}}_i} \in {\mathcal{D}_s}} {{L_y}\left( {{G_y}\left( {{G_f}\left( {{{\mathbf{x}}_i}} \right)} \right),{y_i}} \right)} \\
	         & - \frac{\lambda }{{{n_s} + {n_t}}}\sum\limits_{{{\mathbf{x}}_i} \in {{\mathcal{D}_s} \cup {\mathcal{D}_t}}} {{L_d}\left( {{G_d}\left( {{G_f}\left( {{{\mathbf{x}}_i}} \right)} \right),{d_i}} \right)} 
\end{aligned}         
\end{equation}
where $d_i$ is the domain label of $\mathbf{x}_i$, and $\lambda$ is a hyper-parameter to trade off the two objectives $L_y$ and $L_d$.
After training convergence, the learned parameters $\hat\theta_f$, $\hat\theta_y$, $\hat\theta_d$ will deliver a saddle point of functional~\eqref{eqn:GRL} in the minimax optimization: 
\begin{equation}\label{eqn:param1}
\begin{gathered}
     (\hat\theta_f, \hat\theta_y) =  \mathop {\arg\min }\limits_{{\theta _f},{\theta _y}} C_0 \left( {{\theta _f},{\theta _y},{\theta _d}} \right), \\
     (\hat\theta_d) =  \mathop {\arg\max }\limits_{{\theta_d}} C_0 \left( {{\theta _f},{\theta _y},{\theta _d}} \right).
\end{gathered}
\end{equation}
Domain adversarial neural network has been widely applied to standard domain adaptation where the source domain label space and target domain label space are exactly the same, $\mathcal{C}_s = \mathcal{C}_t$, proving powerful for computer vision problems.

\begin{figure}[tbp]
  \centering
  \includegraphics[width=0.86\textwidth]{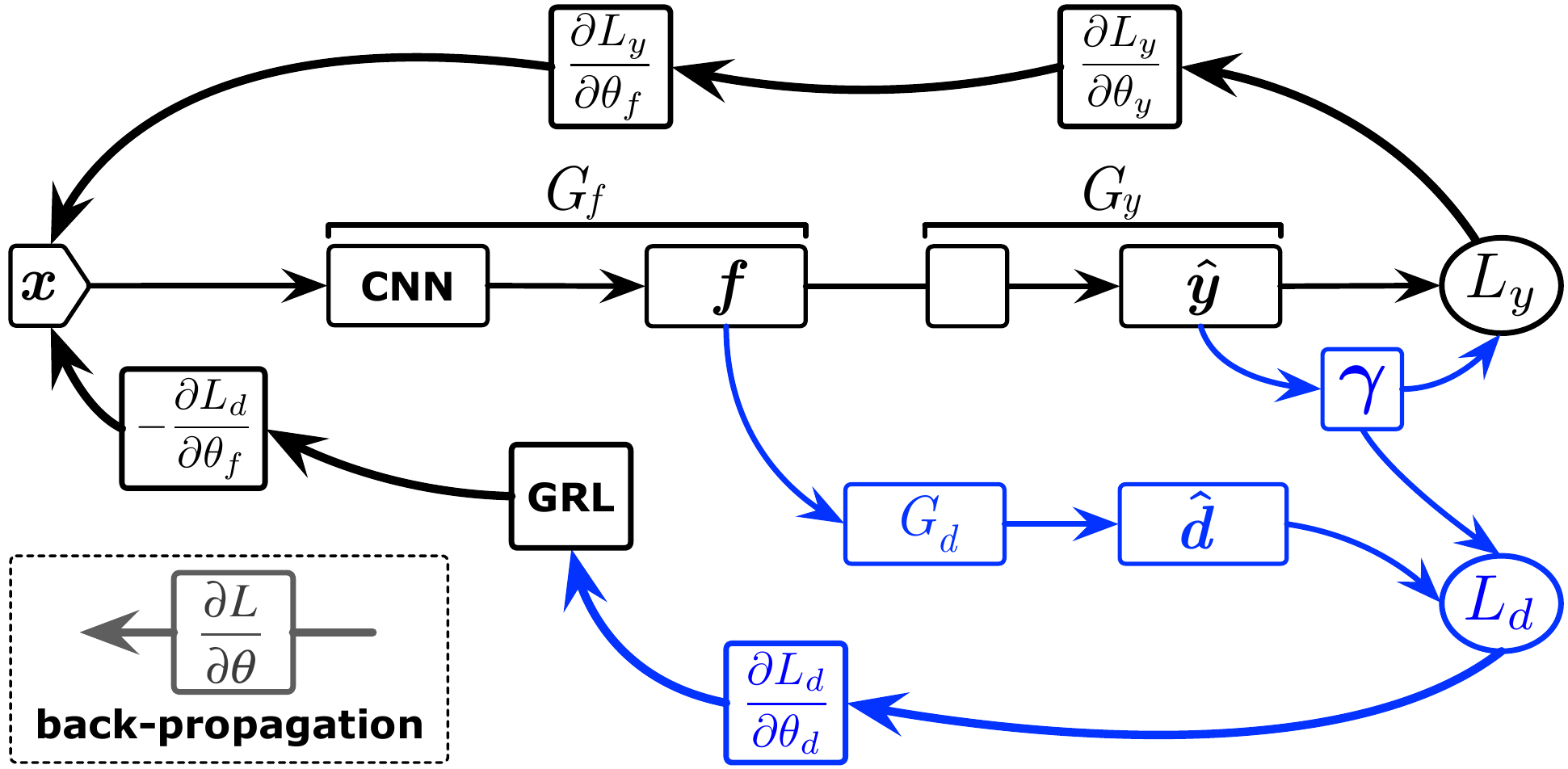}
  \caption{The architecture of Partial Adversarial Domain Adaptation (PADA), where $\mathbf{f}$ is the extracted deep features, ${\hat{\mathbf{y}}}$ is the data label prediction (softmax probability), and ${\hat{\mathbf{d}}}$ is the domain label prediction; $G_f$ is the feature extractor, $G_y$ and $L_y$ are the source classifier and its loss, $G_d$ and $L_d$ are the domain discriminator and its loss; ${\bm\gamma}$ is the class weights averaged over the label predictions of target data. GRL stands for Gradient Reversal Layer. The blue part shows the partial adversarial domain discriminator with weighting mechanism newly designed in this paper. \emph{Best viewed in color.}}
   \label{fig:PADA}
\end{figure}

\subsection{Partial Adversarial Domain Adaptation}
In partial domain adaptation, the target domain label space is a subspace of the source domain label space, $\mathcal{C}_t \subset \mathcal{C}_s$. Thus, aligning the whole source domain distribution $p$ and target domain distribution $q$ will cause negative transfer since the target domain is also forced to match the \emph{outlier} label space $\mathcal{C}_s \backslash \mathcal{C}_t$ in the source domain. And the larger the outlier label space $\mathcal{C}_s \backslash \mathcal{C}_t$ compared to the target label space $\mathcal{C}_t$, the severer the negative transfer will be. Note that $\left| {{\mathcal{C}_t}} \right| \ll \left| {{\mathcal{C}_s}\backslash {\mathcal{C}_t}} \right|$ is a natural scenario of partial domain adaptation in real world applications because we usually need to transfer from very large-scale source dataset (e.g. ImageNet) to relatively small target dataset. 
Thus, when performing adversarial domain adaptation to mitigate \textbf{negative transfer}, we should reduce or even eliminate the negative influence of the outlier source classes as well as the associated source labeled data in $\mathcal{C}_s \backslash \mathcal{C}_t$. 

In this paper, we propose a novel approach, Partial Adversarial Domain Adaptation (\textbf{PADA}), to address the aforementioned challenges. The key idea is to down-weigh the contribution of the source data within the outlier source label space $\mathcal{C}_s \backslash \mathcal{C}_t$ to the training of both the source classifier and the domain adversarial network. This idea is implemented in a novel architecture shown in Fig.~\ref{fig:PADA}. For now the technical problem is reduced to finding some metrics that have large difference between the source outlier classes and the target classes, in order to discriminate the source data belonging to the outlier label space and the target label space. Fortunately, we observe that the output of the source classifier  ${\hat{\bf{y}}}_i = G_y(\mathbf{x}_i)$ to each data point $\mathbf{x}_i$ gives a probability distribution over the source label space $\mathcal{C}_s$. This distribution well characterizes the probability of assigning $\mathbf{x}_i$ to each of the $|\mathcal{C}_s|$ classes. Since the source outlier label space and target label space are disjoint, the target data should be significantly dissimilar to the source data in the outlier label space. Hence, the probabilities of assigning the target data to the source outlier classes, i.e. $y_i^k, {\bf x}_i \in \mathcal{D}_t, k \in \mathcal{C}_s \backslash \mathcal{C}_t$, should be sufficiently small. It is possible that the source classifier can make a few mistakes on some target data and assign large probabilities to false classes or even to outlier classes. To eliminate the influence of such few mistakes, we propose to average the label predictions ${\hat{\bf{y}}}_i$ on all target data. 
Hence, the weight indicating the contribution of each source class to the training can be calculated as follows
\begin{equation}\label{eqn:weight}
    {\bm\gamma} = \frac{1}{n_t}\sum\limits_{i=1}^{n_t}{{\hat{\bf{y}}}_i},
\end{equation}
where ${\bm\gamma}$ is a $|\mathcal{C}_s|$-dimensional weight vector quantifying the contribution of each source class. Specifically, since the target data are not belonging to the source outlier label space, the corresponding weight for source outlier labels ${\gamma}_k, k \in \mathcal{C}_s \backslash \mathcal{C}_t$ will be significantly smaller than the weight for target labels ${\gamma}_k, k \in \mathcal{C}_t$. In practice, it is possible that some of the weights are very small, since by definition, $\sum\nolimits_{k = 1}^{\left| {{\mathcal{C}_s}} \right|} {{\gamma _k}}  = 1$. Thus, we normalize the weight ${\bm\gamma}$ by dividing its largest element, i.e. ${\bm\gamma}  \leftarrow {\bm\gamma} /\max \left( {\bm\gamma}  \right)$. 

We enable partial adversarial domain adaptation by down-weighing the contributions of all source data belonging to the outlier source label space $\mathcal{C}_s \backslash \mathcal{C}_t$. This is achieved by applying the class weight vector ${\bm\gamma}$ to both the source label classifier and the partial adversarial domain discriminator over the source domain data. The objective of the Partial Adversarial Domain Adaptation (\textbf{PADA}) is
\begin{equation}\label{eqn:MultiA}
\begin{aligned}
  C\left( {{\theta _f},{\theta _y},\theta _d} \right) &= \frac{1}{{{n_s}}}\sum\limits_{{\mathbf{x}_i} \in {\mathcal{D}_s}} {\gamma_{y_i}{L_y}\left( {{G_y}\left( {{G_f}\left( {{\mathbf{x}_i}} \right)} \right)}, y_i \right)}  \\
  & - \frac{\lambda}{{n_s}} \sum\limits_{{\mathbf{x}_i} \in {\mathcal{D}_s}} {\gamma_{y_i}L_d\left( {G_d\left( {{G_f}\left( {{{\mathbf{x}}_i}} \right)} \right),d_i} \right)} \\
  & - \frac{\lambda}{{n_t}} \sum\limits_{{\mathbf{x}_i} \in {\mathcal{D}_t}} {L_d\left( {G_d\left( {{G_f}\left( {{{\mathbf{x}}_i}} \right)} \right),d_i} \right)} 
\end{aligned}
\end{equation}  
where $y_i$ is the ground truth label of source point $\mathbf{x}_i$ while $\gamma_{y_i}$ is the corresponding class weight, and $\lambda$ is a hyper-parameter that trade-offs the source label classifier and the partial adversarial domain discriminator in the optimization problem.

The optimization problem finds the optimal parameters ${\hat\theta_f}$, ${\hat\theta_y}$ and ${\hat\theta_d}$ by
\begin{equation}\label{eqn:parameter1}
\begin{gathered}
     ({\hat\theta_f}, {\hat\theta_y}) =  \mathop {\arg\min }\limits_{{\theta _f},{\theta _y}} C\left( {{\theta _f},{\theta _y},\theta _d} \right), \\
     ({\hat\theta_d}) =  \mathop {\arg\max }\limits_{\theta_d} C\left( {\theta _f},{\theta _y},{\theta _d} \right).
\end{gathered}
\end{equation}
Note that the proposed PADA approach as in Equation~\eqref{eqn:parameter1} successfully enables partial domain adaptation, which simultaneously mitigates negative transfer by filtering out outlier source classes $\mathcal{C}_s \backslash \mathcal{C}_t$, and promotes positive transfer by maximally matching the data distributions $p_{\mathcal{C}_t}$ and $q$ in the shared label space $\mathcal{C}_t$.

\section{Experiments}
We conduct experiments on three datasets to evaluate our partial adversarial domain adaptation approach against several state-of-the-art deep transfer learning methods. Codes and datasets is available at \url{https://github.com/thuml/PADA}.

\subsection{Setup}
The evaluation is conducted on three standard domain adaptation datasets: Office-31, Office-Home and ImageNet-Caltech.

\textbf{Office-31}~\cite{cite:ECCV10Office} is a most widely-used dataset for visual domain adaptation, with 4,652
images and 31 categories from three distinct domains: \textit{Amazon} (\textbf{A}), which contains images
downloaded from amazon.com, \textit{Webcam} (\textbf{W}) and \textit{DSLR} (\textbf{D}), which contain images respectively taken by web camera and digital SLR camera. We denote the three domains as \textbf{A}, \textbf{W} and \textbf{D}. We use the ten categories shared by \textit{Office-31} and \textit{Caltech-256} and select images of these ten categories in each domain of \textit{Office-31} as target domains. We evaluate all methods across
six partial domain adaptation tasks \textbf{A} $\rightarrow$ \textbf{W}, \textbf{D} $\rightarrow$ \textbf{W}, \textbf{W} $\rightarrow$ \textbf{D}, \textbf{A} $\rightarrow$ \textbf{D}, \textbf{D} $\rightarrow$ \textbf{A} and \textbf{W} $\rightarrow$ \textbf{A}. Note that each source domain here contains 31 categories and each target domain here contains 10 categories.

\begin{figure}[h]
  \centering
  \includegraphics[width=0.74\columnwidth]{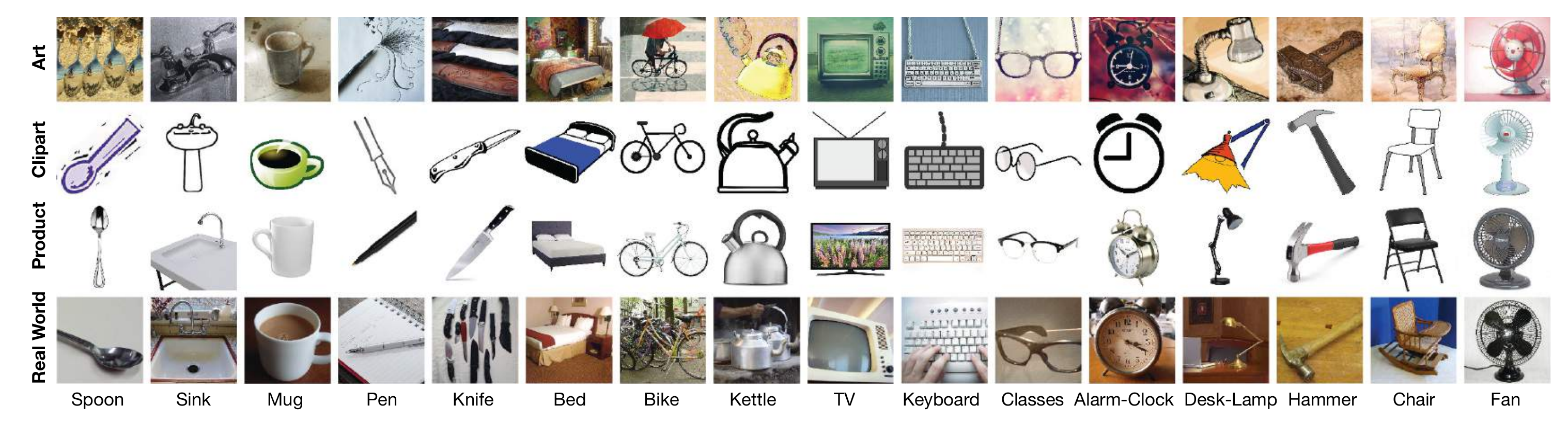}
  \caption{Example images of the Office-Home dataset.}
  \label{fig:data}
\end{figure}

\textbf{Office-Home}~\cite{cite:CVPR17OfficeHome} was released recently as a more challenging domain adaptation dataset, crawled through several search engines and online image directories, as shown in Fig.~\ref{fig:data}. It consists of 4 different domains: Artistic images (\textbf{Ar}), Clipart images (\textbf{Cl}), Product images (\textbf{Pr}) and Real-World images (\textbf{Rw}). For each domain, the dataset contains images from 65 object categories. In each transfer task, when a domain is used as source domain, we use the images from all 65 categories; when a domain is used as target domain, we choose (in alphabetic order) the first 25 categories as target categories and select all images of these 25 categories as target domain. Denoting the four domains as \textbf{Ar}, \textbf{Cl}, \textbf{Pr}, \textbf{Rw}, we can build twelve partial domain adaptation tasks: \textbf{Ar} $\rightarrow$ \textbf{Cl}, \textbf{Ar} $\rightarrow$ \textbf{Pr}, \textbf{Ar} $\rightarrow$ \textbf{Rw}, \textbf{Cl} $\rightarrow$ \textbf{Ar}, \textbf{Cl} $\rightarrow$ \textbf{Pr}, \textbf{Cl} $\rightarrow$ \textbf{Rw}, \textbf{Pr} $\rightarrow$ \textbf{Ar}, \textbf{Pr} $\rightarrow$ \textbf{Cl}, \textbf{Pr} $\rightarrow$ \textbf{Rw}, \textbf{Rw} $\rightarrow$ \textbf{Ar}, \textbf{Rw} $\rightarrow$ \textbf{Cl} and \textbf{Rw} $\rightarrow$ \textbf{Pr}. The transfer tasks from this dataset can test the performance of our method on more visually-dissimilar domains. 

\textbf{ImageNet-Caltech} is constructed from \textit{ImageNet-1K}~\cite{cite:ILSVRC15} and \textit{Caltech-256}. They share 84 common classes, and thus we form two partial domain adaptation tasks: \textbf{ImageNet-1K} $\rightarrow$ \textbf{Caltech-84} and \textbf{Caltech-256} $\rightarrow$ \textbf{ImageNet-84}. To prevent the effect of the pre-trained model on ImageNet, we adopt the validation set when ImageNet is used as target domain and adopt the training set when ImageNet is used as source domain. This setting represents the performance on partial domain adaptation tasks with large number of classes.

\textbf{VisDA2017} has two domains where one consists of synthetic 2D renderings of 3D models and the other consists of photo-realistic images or real images. They have 12 classes in common. We choose (in alphabetic order) the first 6 categories as target categories and select all images of these 6 categories as target domain and construct two domain adaptation tasks: \textbf{Real-12} $\rightarrow$ \textbf{Synthetic-6} and \textbf{Synthetic-12} $\rightarrow$ \textbf{Real-6}. In this dataset, both domains have large number of images, which validates the efficiency of PADA on large-scale dataset.

We compare the performance of \textbf{PADA} with state-of-the-art transfer learning and deep learning methods: \textbf{ResNet-50}~\cite{cite:CVPR16DRL}, Deep Adaptation Network (\textbf{DAN})~\cite{cite:ICML15DAN}, Residual Transfer Networks (\textbf{RTN})~\cite{cite:NIPS16RTN}, Domain Adversarial Neural Network (\textbf{DANN})~\cite{cite:ICML15RevGrad}, Adversarial Discriminative Domain Adaptation (\textbf{ADDA}) \cite{cite:CVPR17ADDA} and Joint Adaptation Network (\textbf{JAN})~\cite{cite:ICML17JAN}. In order to go deeper with the efficacy of the proposed partial adversarial mechanism, we perform ablation study by evaluating two variants of \textbf{PADA}: (1) \textbf{PADA-classifier} is the variant without the weight on the source classifier; (2) \textbf{PADA-adversarial} is the variant without the weight on the partial adversarial domain discriminator.

We follow standard evaluation protocols and use all labeled source data and all unlabeled target data for unsupervised domain adaptation~\cite{cite:ECCV10Office,cite:ICML15DAN}. We compare the average classification accuracy of each partial domain adaptation task using three random experiments. For MMD-based methods (DAN and RTN), we use Gaussian kernel with bandwidth set to median pairwise squared distances on training data, i.e. the median heuristic~\cite{cite:NIPS12MKMMD}. For all methods, we use importance-weighted cross-validation~\cite{cite:JMLR07IWCV} on labeled source data and unlabeled target data to select their hyper-parameters.

We implement all deep methods in \textbf{PyTorch}, and fine-tune from PyTorch-provided models of ResNet-50~\cite{cite:NIPS12CNN} pre-trained on ImageNet. We add a bottleneck layer between the $res5c$ and $fc$ layers as DANN \cite{cite:ICML15RevGrad} except for the transfer task \textbf{ImageNet (1000 classes)} $\rightarrow$ \textbf{Caltech (84 classes)}, since the pre-trained model is trained on ImageNet dataset and it can fully exploit the advantage of pre-trained model with the original network parameters. For PADA, we fine-tune all the feature layers and train the bottleneck layer, the classifier layer and the partial adversarial domain discriminator though back-propagation. Since these new layers are trained from scratch, we set their learning rate to be 10 times that of the other layers. We use mini-batch stochastic gradient descent (SGD) with momentum of 0.9 and the learning rate strategy implemented in DANN~\cite{cite:ICML15RevGrad}: the learning rate is adjusted during SGD using $\eta_p = \frac{\eta_0}{{(1+\alpha p)}^\gamma}$, where $p$ is the training progress changing from 0 to 1,  $\eta_0$, while $\alpha$ and $\gamma$ are optimized with importance-weighted cross-validation~\cite{cite:JMLR07IWCV} on one task of a dataset and fixed for all the other tasks of this dataset. As \textbf{PADA} works stably across different tasks, the penalty of adversarial networks is increased progressively from $0$ to $1$ as DANN \cite{cite:ICML15RevGrad}. 

\begin{table}[bp]
    \centering 
    \caption{Accuracy of partial domain adaptation tasks on \emph{Office-Home} (ResNet-50)}
    \label{table:accuracy_officehome}
    \resizebox{\textwidth}{!}{%
    \begin{tabular}{|c|cccccccccccc|c|}
    	\hline
        \multirow{2}{30pt}{\centering Method} & \multicolumn{13}{c|}{Office-Home} \\
        \cline{2-14}
        & \textbf{Ar}$\rightarrow$\textbf{Cl} & \textbf{Ar}$\rightarrow$\textbf{Pr} & \textbf{Ar}$\rightarrow$\textbf{Rw} & \textbf{Cl}$\rightarrow$\textbf{Ar} & \textbf{Cl}$\rightarrow$\textbf{Pr} & \textbf{Cl}$\rightarrow$\textbf{Rw} & \textbf{Pr}$\rightarrow$\textbf{Ar} & \textbf{Pr}$\rightarrow$\textbf{Cl} & \textbf{Pr}$\rightarrow$\textbf{Rw} & \textbf{Rw}$\rightarrow$\textbf{Ar} & \textbf{Rw}$\rightarrow$\textbf{Cl} & \textbf{Rw}$\rightarrow$\textbf{Pr} & Avg \\
        \hline
        ResNet~\cite{cite:NIPS12CNN} & 38.57 & 60.78 & 75.21 & 39.94 & 48.12 & 52.90 & 49.68 & 30.91 & 70.79 & 65.38 & 41.79 & 70.42 & 53.71 \\
        DAN~\cite{cite:ICML15DAN} & 44.36 & 61.79 & 74.49 & 41.78 & 45.21 & 54.11 & 46.92 & 38.14 & 68.42 & 64.37 & 45.37 & 68.85 & 54.48 \\
        DANN~\cite{cite:ICML15RevGrad} & 44.89 & 54.06 & 68.97 & 36.27 & 34.34 & 45.22 & 44.08 & 38.03 & 68.69 & 52.98 & 34.68 & 46.50 & 47.39 \\
        RTN~\cite{cite:NIPS16RTN} & 49.37 & 64.33 & 76.19 & 47.56 & 51.74 & 57.67 & 50.38 & 41.45 & 75.53 & 70.17 & 51.82 & 74.78 & 59.25 \\
        \hline
        PADA-classifier & 47.45 & 58.15 & 74.32 & 43.62 & 37.93 & 51.91 & 48.21 & 41.67 & 71.62 & 67.13 & 52.98 & 71.60 & 55.55 \\
        PADA-adversarial & 47.10 & 47.54 & 67.53 & 41.32 & 39.72 & 52.70 & 43.07 & 35.94 & 70.51 & 61.80 & 48.24 & 70.08 & 52.13 \\ 
        PADA & \textbf{51.95} & \textbf{67} & \textbf{78.74} & \textbf{52.16} & \textbf{53.78} & \textbf{59.03} & \textbf{52.61} & \textbf{43.22} & \textbf{78.79} & \textbf{73.73} & \textbf{56.6} & \textbf{77.09} & \textbf{62.06} \\ 
    	\hline
    \end{tabular}%
    }
\end{table}

\begin{table}[tbp]
    \addtolength{\tabcolsep}{6pt}
    \centering 
    \caption{Accuracy of partial domain adaptation tasks on \emph{Office-31} (ResNet-50)}
    \label{table:accuracy_officeic}
    \resizebox{\textwidth}{!}{%
    \begin{tabular}{|c|cccccc|c|cc|c|}
    	\hline
        \multirow{2}{30pt}{\centering Method} &  \multicolumn{7}{c|}{Office-31} \\
        \cline{2-8}
        & A $\rightarrow$ W & D $\rightarrow$ W & W $\rightarrow$ D & A $\rightarrow$ D & D $\rightarrow$ A & W $\rightarrow$ A & Avg \\
        \hline
        ResNet~\cite{cite:NIPS12CNN} & 54.52 & 94.57 & 94.27 & 65.61 & 73.17 & 71.71 & 75.64\\
        DAN~\cite{cite:ICML15DAN} & 46.44 & 53.56 & 58.60 & 42.68 & 65.66 & 65.34 & 55.38 \\
        DANN~\cite{cite:ICML15RevGrad} & 41.35 & 46.78 & 38.85 & 41.36 & 41.34 & 44.68 & 42.39  \\
        ADDA~\cite{cite:CVPR17ADDA} & 43.65 & 46.48 & 40.12 & 43.66 & 42.76 & 45.95 & 43.77 \\
        RTN~\cite{cite:NIPS16RTN} & 75.25 & 97.12 & 98.32 & 66.88 & 85.59 & 85.70 & 84.81 \\
        JAN~\cite{cite:ICML17JAN} & 43.39 & 53.56 & 41.40 & 35.67 & 51.04 & 51.57 & 46.11 \\
        LEL~\cite{cite:NIPS17LEL} & 73.22 & 93.90 & 96.82 & 76.43 & 83.62 & 84.76 & 84.79 \\
        \hline
        PADA-classifier & 83.12 & 99.32 & 100 & 80.16 & 90.13 & 92.34 & 90.85  \\
        PADA-adversarial & 65.76 & 97.29 & 97.45 & 77.07 & 87.27 & 87.37 & 85.37  \\ 
        PADA & \textbf{86.54} & \textbf{99.32} & \textbf{100} & \textbf{82.17} & \textbf{92.69} & \textbf{95.41} & \textbf{92.69} \\ 
    	\hline
    \end{tabular}%
    }
\end{table}

\subsection{Results}

The classification accuracy results of partial domain adaptation on the twelve tasks of \textit{Office-Home}, the six tasks of \textit{Office-31}, and the two tasks of \textit{ImageNet-Caltech} are shown in Tables~\ref{table:accuracy_officehome}, \ref{table:accuracy_officeic}, and \ref{table:accuracy_visda}, respectively. PADA outperforms all comparison methods on all the tasks. In particular, PADA substantially improves the average accuracy by huge margins on \textit{Office-31} of small domain gaps, e.g. \textbf{Amazon} $\rightarrow$ \textbf{Webcam}, and on \textit{Office-Home} of large domain gaps, e.g. \textbf{Clipart} $\rightarrow$ \textbf{Real World}. It is inspiring that PADA achieves considerable accuracy gains on \textbf{ImageNet}$\rightarrow$\textbf{Caltech} with large-scale source domain and target domain. These consistent results suggest that PADA can learn transferable features and classifiers for partial domain adaptation on all the partial domain adaptation tasks, varying by the sizes of the source and target domains and the gaps between the source and target domains. 

Previous deep domain adaptation methods, including those based on adversarial networks (e.g. DANN) and those based on MMD (e.g. DAN), perform worse than standard ResNet on most of the tasks, showing the undesirable influence of the \textbf{negative transfer} effect. 
Adversarial-network based methods try to learn deep features that deceive the domain discriminator, while MMD based methods align the source and target feature distributions. Both mechanisms will mix up the whole source and target domains in the feature space, since they aim to match all classes of source domain to target domain. But there are classes in the source domain that do not exist in the target domain, a.k.a. outlier source classes. This explains their weak performance for partial domain adaptation.
Not surprisingly, PADA outperforms all the comparison methods by large margins, indicating that PADA can effectively avoid negative transfer by eliminating the influence of outlier source classes irrelevant to the target domain. 

Methods based on domain-adversarial networks perform worse than MMD-based methods. Since adversarial-network based methods try to confuse a nonlinear domain discriminator, it has more power to match source and target domains and is more vulnerable to the outlier source classes than MMD based methods. 
Although PADA is also based on adversarial networks, it establishes a partial adversarial domain discriminator,  which down-weighs the source data in the outlier source classes to eliminate the negative influence of the outlier source classes and meanwhile enhances the positive influence of shared classes, successfully boosting the performance of partial domain adaptation.

\begin{table}[tbp]
    \centering 
    \caption{Classification accuracy on \emph{ImageNet-Caltech} and \emph{VisDA2017} (ResNet-50)}
    \label{table:accuracy_visda}
    \resizebox{\textwidth}{!}{%
    \begin{tabular}{|c|ccc|ccc|}
    	\hline
        \multirow{2}{30pt}{\centering Method} &  \multicolumn{3}{c|}{ImageNet-Caltech} & \multicolumn{3}{c|}{VisDA2017}\\
        \cline{2-7}
        & ImageNet $\rightarrow$ Caltech & Caltech $\rightarrow$ ImageNet & Avg & Real $\rightarrow$ Synthetic & Synthetic $\rightarrow$ Real & Avg \\
        \hline
        ResNet~\cite{cite:NIPS12CNN} & 71.65 & 66.14 & 68.90 & 64.28 & 45.26 & 54.77 \\
        DAN~\cite{cite:ICML15DAN} & 71.57 & 66.48 & 69.03 & 68.35 & 47.60 & 57.98 \\
        DANN~\cite{cite:ICML15RevGrad} & 68.67 & 52.97 & 60.82 & 73.84 & 51.01 & 62.43  \\
        RTN~\cite{cite:NIPS16RTN} & 72.24 & 68.33 & 70.29 &  72.93& 50.04 & 61.49 \\
        \hline
        PADA & \textbf{75.03} & \textbf{70.48} & \textbf{72.76} & 76.50 & 53.53 &  65.01 \\ 
    	\hline
    \end{tabular}%
    }
\end{table}

Among previous deep domain adaptation methods, RTN is the only approach that generally performs better than ResNet-50. The remedy of RTN is to introduce entropy minimization criterion which can increase the contributions of target data and thus implicitly avoid the impact of outlier source data to some degree. Unlike RTN, PADA does not use the entropy minimization criterion. But we observe that PADA outperforms RTN in all the partial domain adaptation tasks, proving that RTN also suffers from the negative transfer effect and even the residual branch of RTN, originally designed for large domain gap, cannot bridge the large discrepancy between source and target caused by different label spaces. These results validate that our partial adversarial mechanism is versatile enough to jointly promote positive transfer from relevant source domain data (belonging to target label space) and circumvent negative transfer from irrelevant source domain data (belonging to outlier source label space).

We perform ablation study of PADA by comparing the PADA variants in Tables~\ref{table:accuracy_officehome} and ~\ref{table:accuracy_officeic}. We can make some insightful observations from the results. \textbf{(1)} PADA outperforms PADA-classifier, proving that using weighting mechanism on the source classifier can reduce the influence of the source data in the outlier classes and focus the source classifier more on the source data belonging to the target label space. \textbf{(2)} PADA outperforms PADA-adversarial with large margin, which proves that our weighting mechanism on the domain adversarial network can assign small weight on the outlier classes and down-weigh the source data of the outlier classes. In this way, PADA is able to avoid matching the whole source domain and target domain and boost performance of partial domain adaptation. 

\begin{figure}[!bp]
  \centering
  \subfigure[PADA: \textbf{A}$\rightarrow$\textbf{W}]{
    \includegraphics[width=0.35\textwidth]{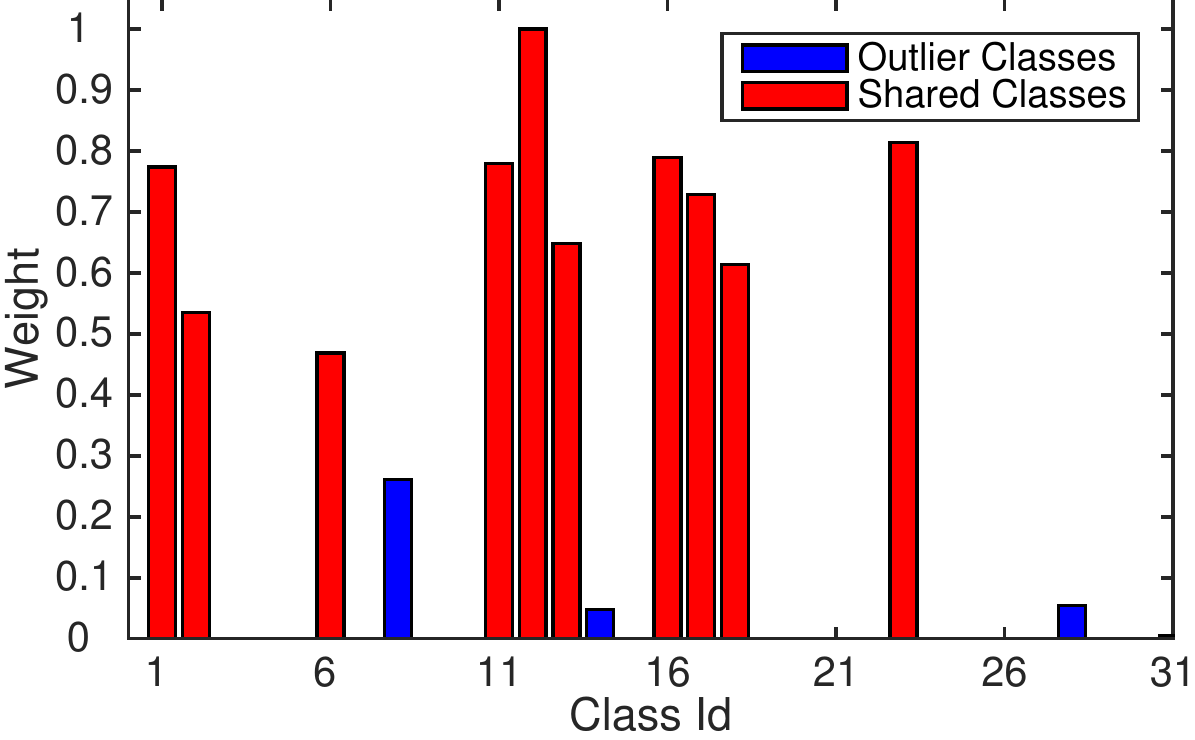}
    \label{fig:weight_of_office}
  }
  \hfil
  \subfigure[PADA: \textbf{ImageNet-1K}$\rightarrow$\textbf{Caltech-84}]{
    \includegraphics[width=0.30\textwidth]{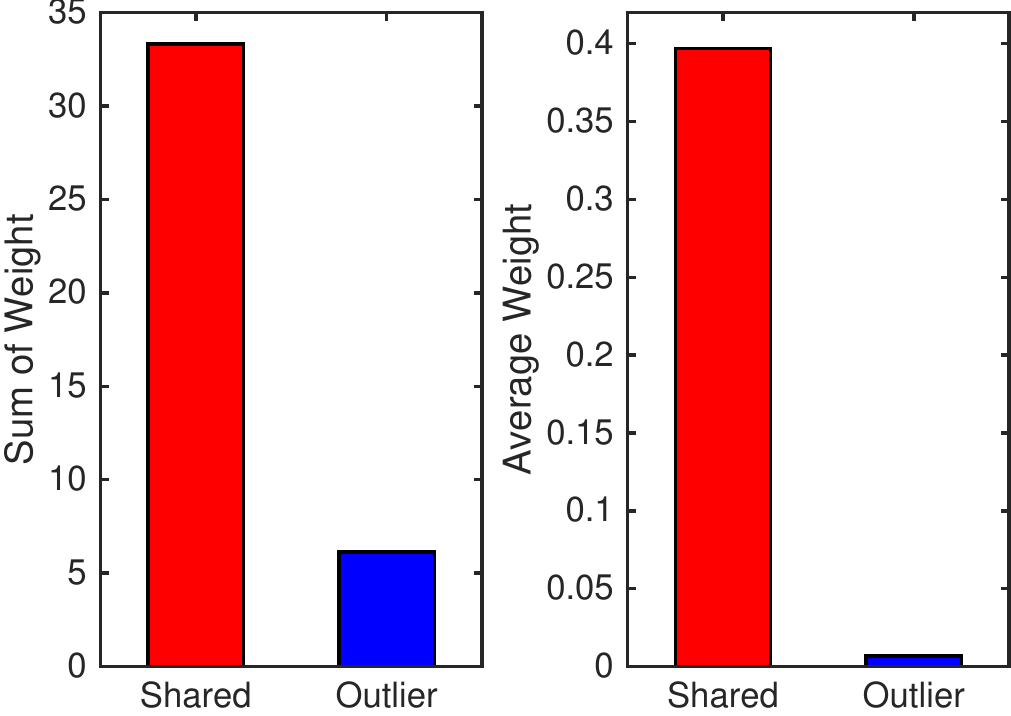}
    \label{fig:weight_of_imagenet}
  }
  \\
  \subfigure[DANN: \textbf{A}$\rightarrow$\textbf{W}]{
    \includegraphics[width=0.35\textwidth]{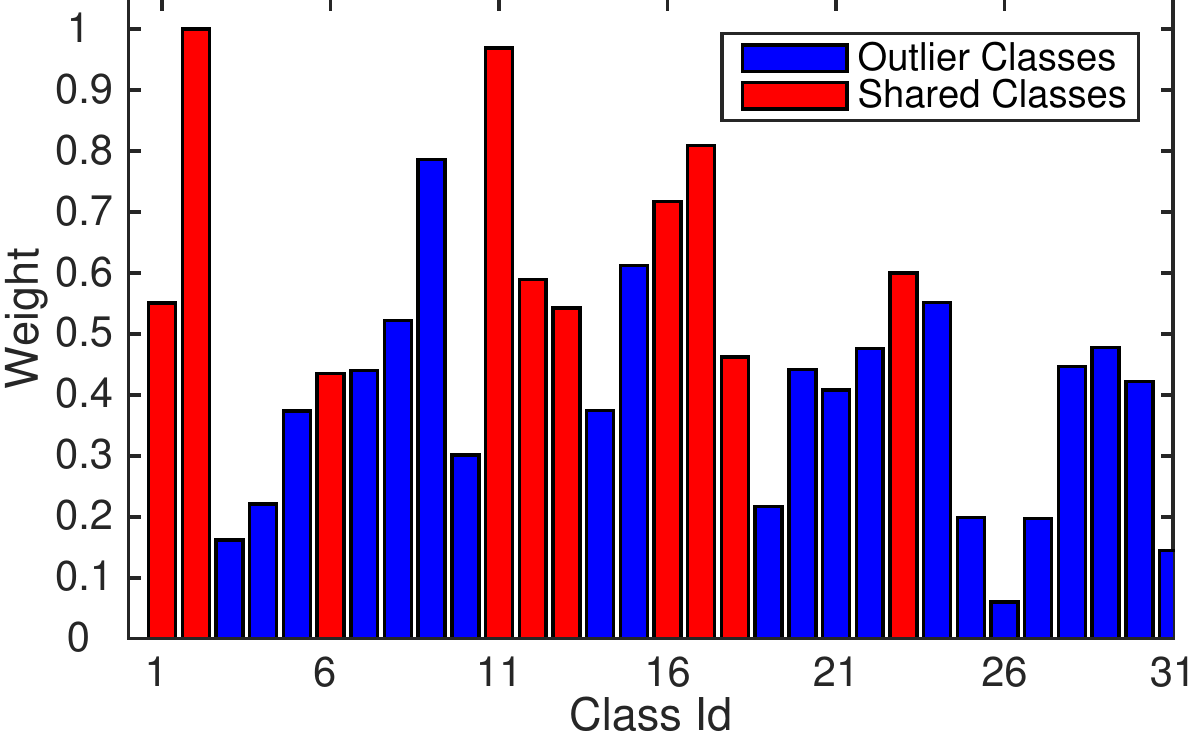}
    \label{fig:weight_of_office_grl}
  }
  \hfil
  \subfigure[DANN: \textbf{ImageNet-1K}$\rightarrow$\textbf{Caltech-84}]{
    \includegraphics[width=0.30\textwidth]{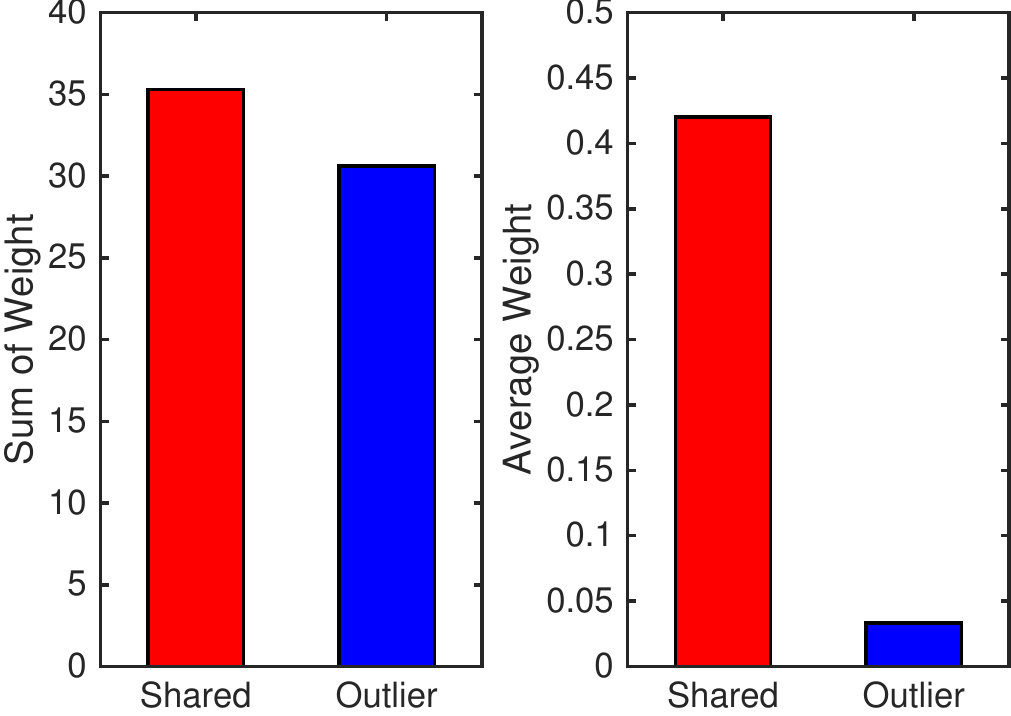}
    \label{fig:weight_of_imagenet_grl}
  }
  \caption{Histograms of class weights learned by PADA and DANN on two typical tasks.}
\end{figure}

\subsection{Empirical Analysis}

\textbf{Statistics of Class Weights:} We illustrate the learned class weight for each class in our weighting mechanism on the task \textbf{A} (31 classes) $\rightarrow$ \textbf{W} (10 classes) in Fig.~\ref{fig:weight_of_office}. Since it is difficult to visualize all the 1000 weights in one plot, we visualize the sum and average of the weights of all classes on the task \textbf{ImageNet-1K} $\rightarrow$ \textbf{Caltech-84} in Fig.~\ref{fig:weight_of_imagenet}. As shown in Fig.~\ref{fig:weight_of_office}, our partial adversarial mechanism assigns much larger weights to the shared classes than to the outlier classes. It is inspiring that and some outlier classes even have nearly zero weights, by carefully observing the plot for task \textbf{A}$\rightarrow$\textbf{W}. Not only for task with small number of classes, our partial adversarial mechanism also works well on dataset with large number of classes, as shown in~\ref{fig:weight_of_imagenet}. In task \textbf{ImageNet-1K} $\rightarrow$ \textbf{Caltech-84}, the average weights of the shared classes are significantly larger than those of the outlier classes. Note that even though the number of the outlier classes is much larger than that of the shared classes, the sum of the weights for shared classes is still larger than the sum of the weights for outlier classes. These results prove that PADA can automatically discriminate the shared and outlier classes and down-weigh the outlier classes data. 

We also show the corresponding class weights learned by DANN on task \textbf{A} (31 classes) $\rightarrow$ \textbf{W} (10 classes) in Fig.~\ref{fig:weight_of_office_grl}, and task \textbf{ImageNet-1K} $\rightarrow$ \textbf{Caltech-84} in Fig.~\ref{fig:weight_of_imagenet_grl}. The results clearly demonstrate that the weights for outlier source classes are still substantially large, which cannot down-weigh these harmful outlier classes in the domain adversarial learning procedure. As a result, these outlier source classes will cause severe negative transfer, as shown in Tables~\ref{table:accuracy_officehome} and \ref{table:accuracy_officeic}.

\begin{figure}[!htbp]
  \centering
  \subfigure[Acc w.r.t \#Target Classes]{
    \includegraphics[width=0.35\textwidth]{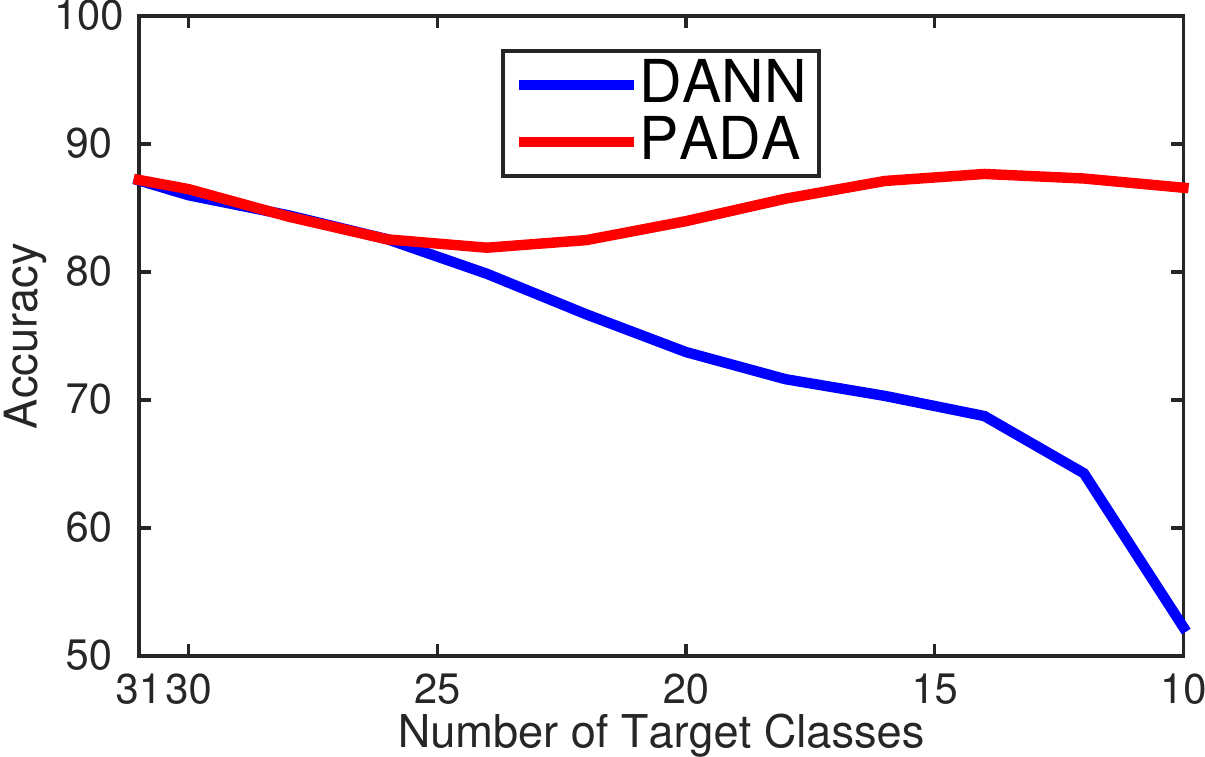}
    \label{fig:accuracy_number}
  }\hfil
  \subfigure[Test Error]{
    \includegraphics[width=0.36\textwidth]{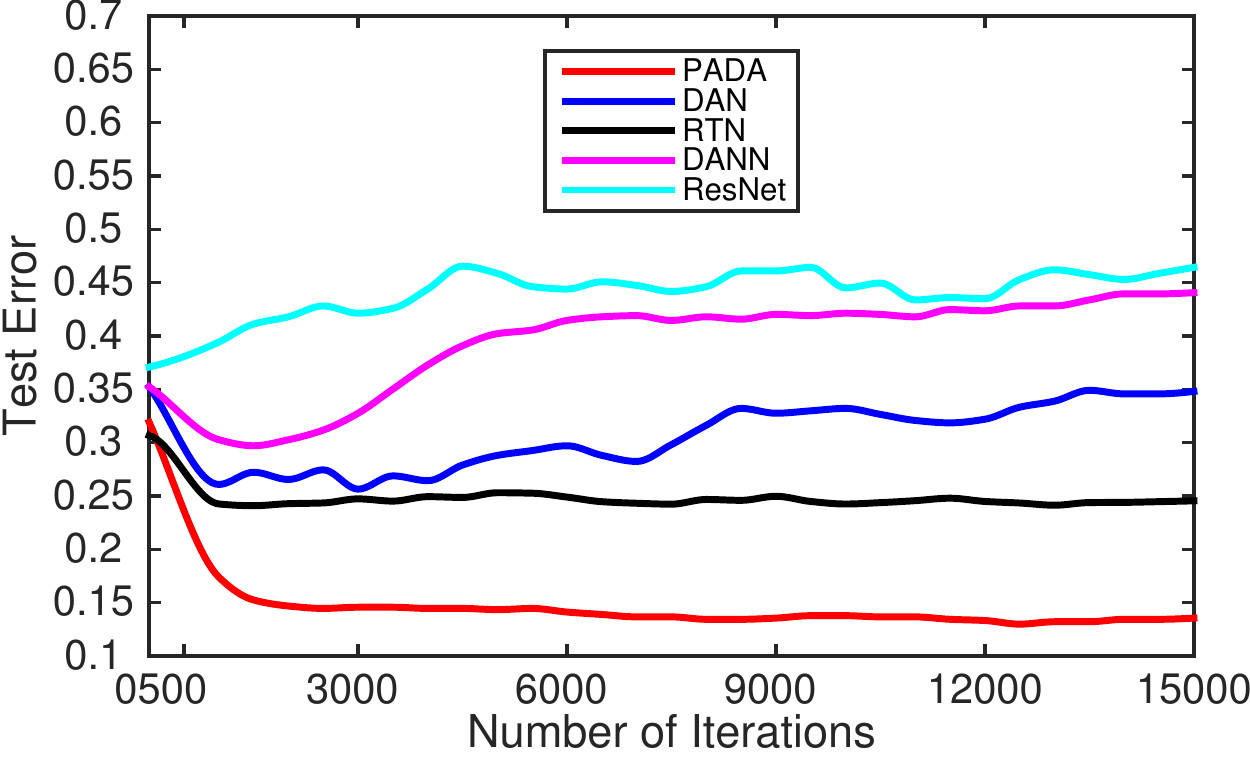}
    \label{fig:validation_error}
  }
  \caption{Analysis: (a) Accuracy by varying \#target domain classes; (b) Target test error.}
\end{figure}

\textbf{Accuracy for Different Numbers of Target Classes:}
We investigate a wider spectrum of partial domain adaptation by varying the number of target classes. Fig.~\ref{fig:accuracy_number} shows that when the number of target classes decreases, the performance of DANN degrades quickly, meaning that negative transfer becomes severer when the domain gap is enlarged. The performance of PADA degenerates when the number of target classes decreases from $31$ to $25$, where negative transfer problem arises while the domain adaptation problem itself is still hard; but it increases when the number of target classes decreases from $25$ to $10$, where the domain adaptation problem itself becomes easier. The margin that PADA outperforms DANN becomes larger when the number of target classes decreases. Note that PADA performs comparably to DANN in standard domain adaptation when the number of target classes is $31$, meaning that the weighting mechanism will not wrongly filter out classes when there are no outlier classes.

\textbf{Convergence Performance:}
We examine the convergence of PADA by studying the test error through training process. As shown in Fig.~\ref{fig:validation_error}, the test errors of ResNet, DAN and DANN are increasing due to negative transfer. Their test errors are not stable as well, attributed to the possibility that the target domain is falsely matched to different parts of the source domain during the training process.  RTN converges very fast depending on the entropy minimization criterion, but it converges to a higher test error. PADA converges fast and stably to the lowest test error, implying that it can be trained efficiently and stably to tackle partial domain adaptation problems. 

\begin{figure}[htbp]
  \centering
  \subfigure[DAN]{
    \includegraphics[width=0.18\textwidth]{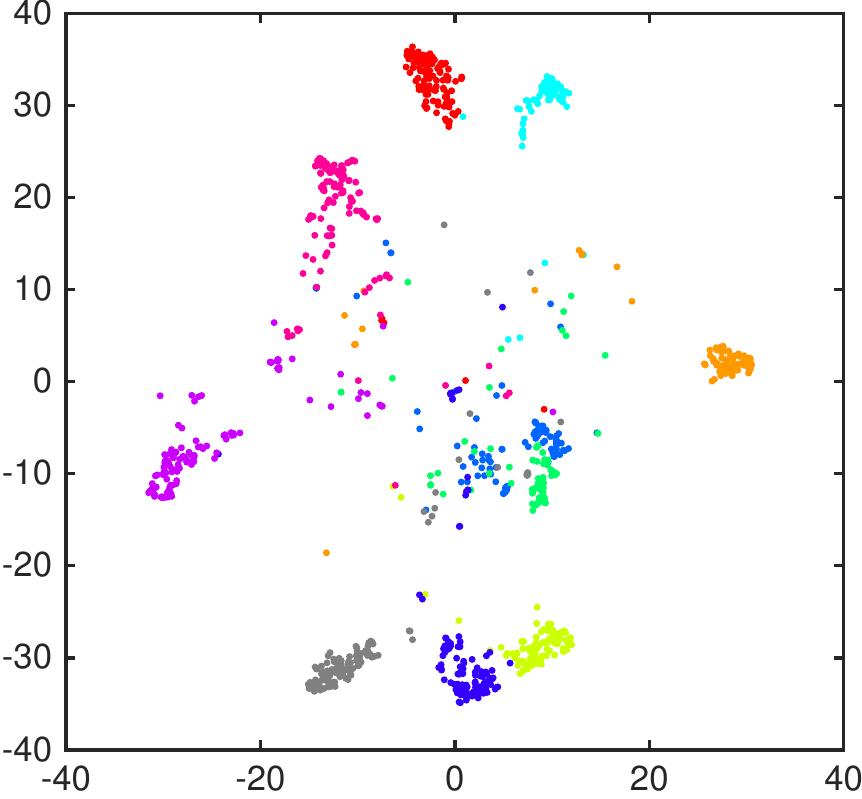}
    \label{fig:dan}
  }\hfil
  \subfigure[DANN]{
    \includegraphics[width=0.18\textwidth]{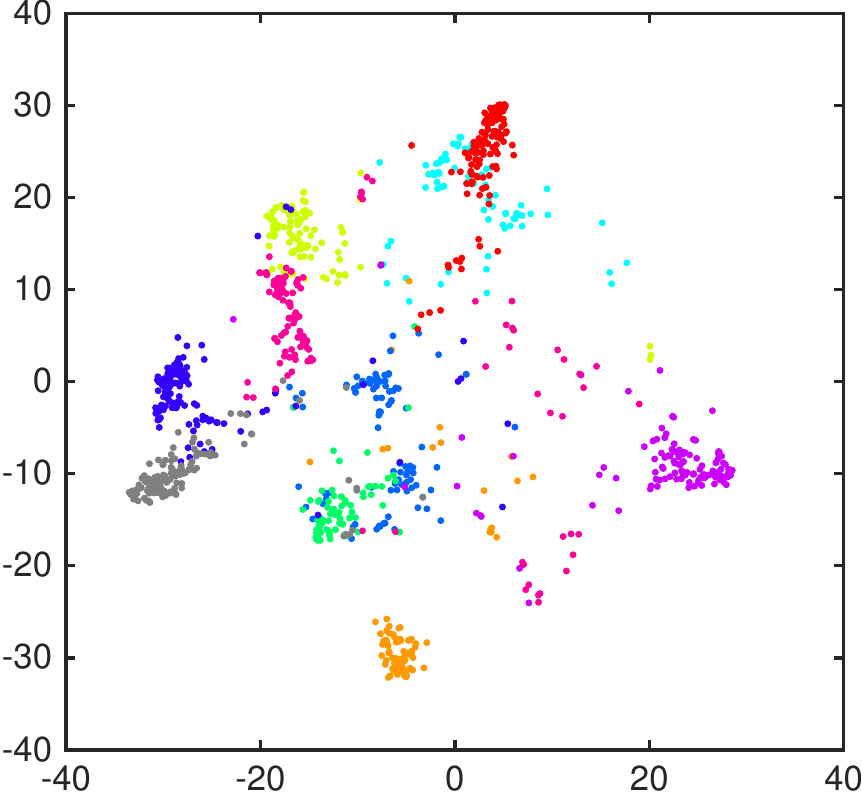}
    \label{fig:grl}
  }\hfil
  \subfigure[RTN]{
    \includegraphics[width=0.18\textwidth]{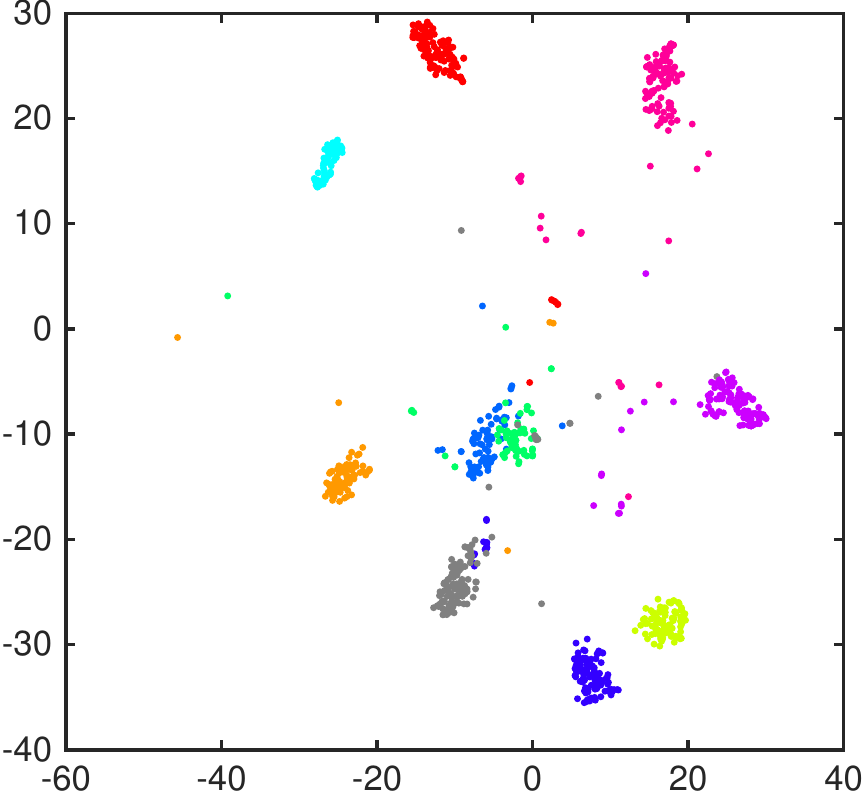}
    \label{fig:rtn}
  }\hfil
  \subfigure[PADA]{
    \includegraphics[width=0.24\textwidth]{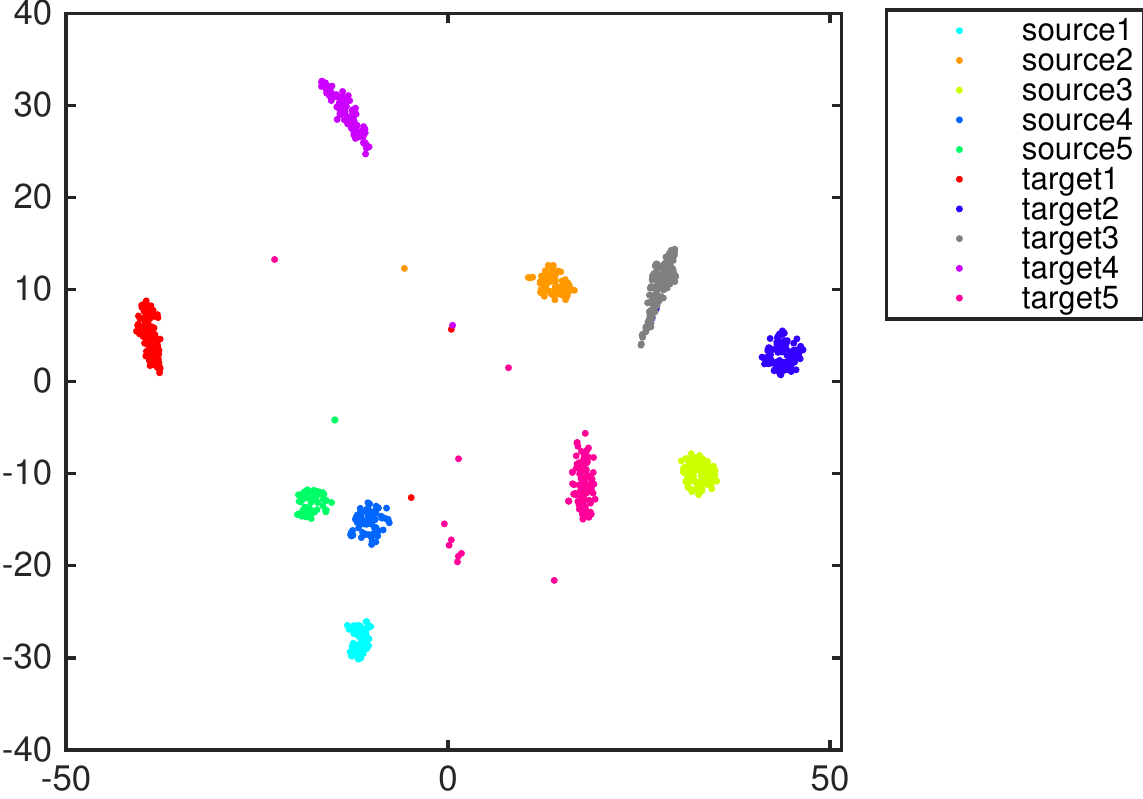}
    \label{fig:pada}
  }
  \caption{The t-SNE visualization of DAN, DANN, RTN, and PADA with class info.}
\end{figure}

\begin{figure}[htbp]
  \centering
  \subfigure[DAN]{
    \includegraphics[width=0.18\textwidth]{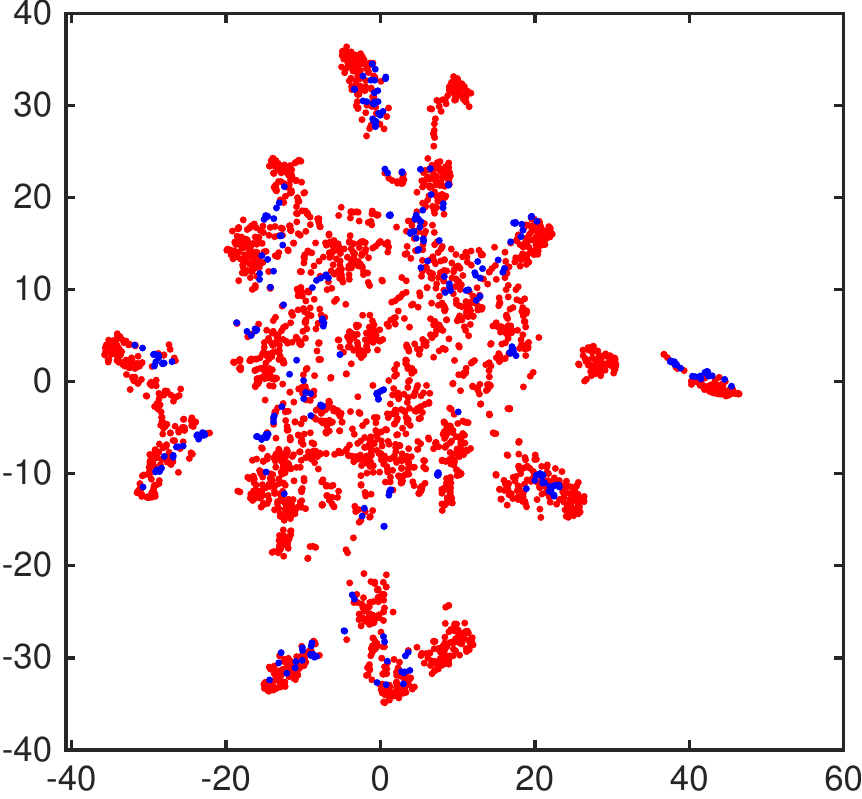}
    \label{fig:dan_st}
  }\hfil
  \subfigure[DANN]{
    \includegraphics[width=0.18\textwidth]{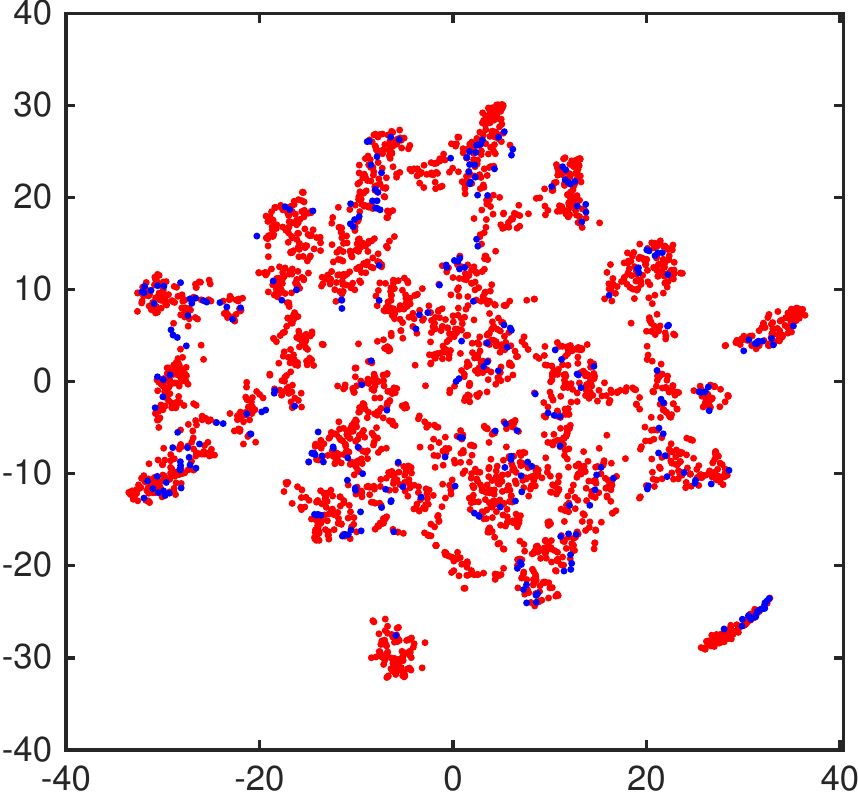}
    \label{fig:grl_st}
  }\hfil
  \subfigure[RTN]{
    \includegraphics[width=0.18\textwidth]{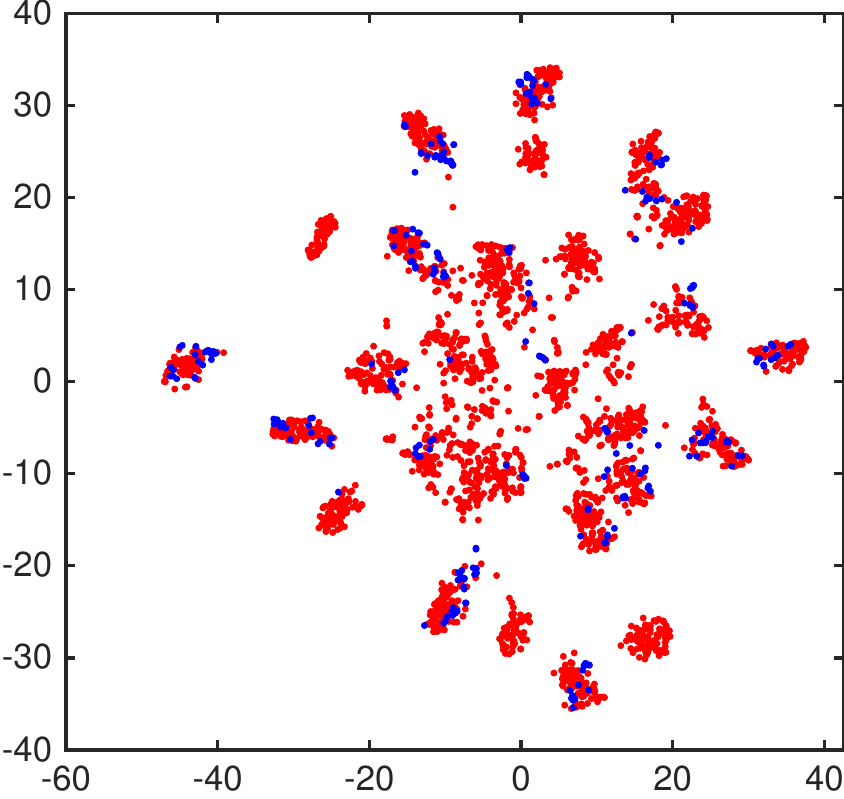}
    \label{fig:rtn_st}
  }\hfil
  \subfigure[PADA]{
    \includegraphics[width=0.25\textwidth]{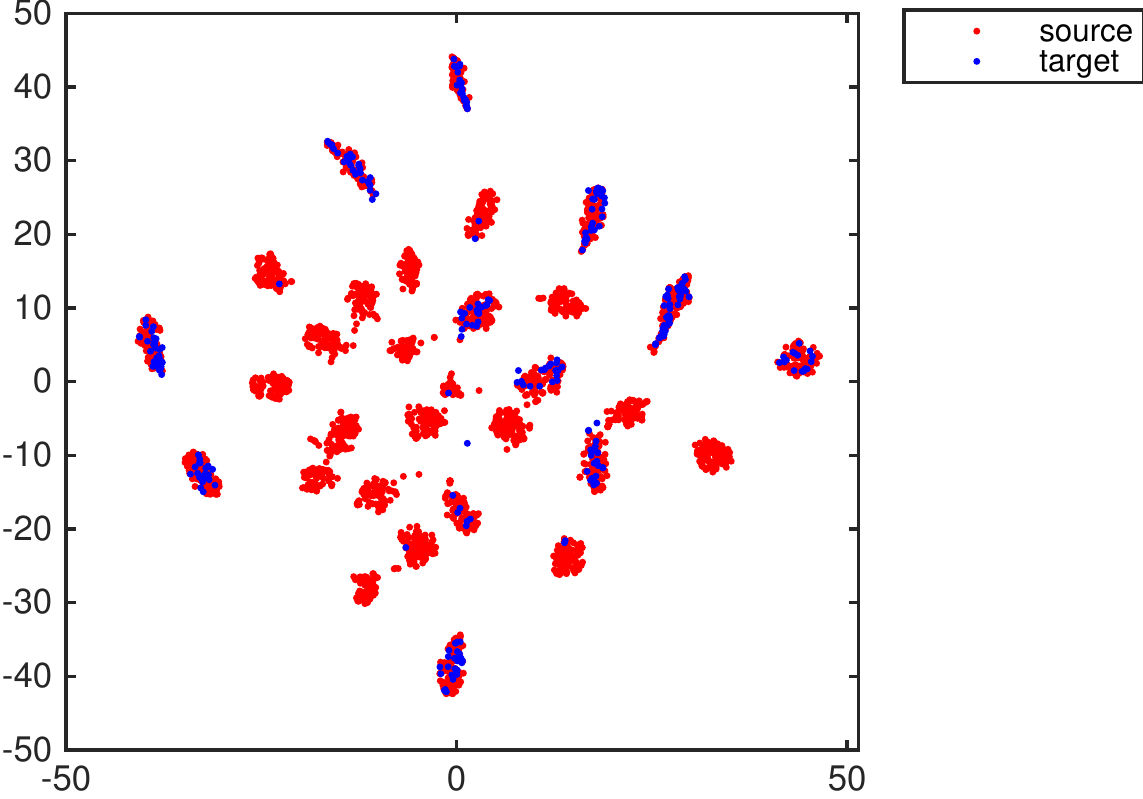}
    \label{fig:pada_st}
  }
  \caption{The t-SNE visualization of DAN, DANN, RTN, and PADA with domain info.}
\end{figure}

\textbf{Feature Visualization:} We visualize the t-SNE embeddings~\cite{cite:ICML14DeCAF} of the bottleneck representations by DAN, DANN, RTN and PADA on the partial domain adaptation task \textbf{A} (31 classes) $\rightarrow$ \textbf{W} (10 classes) in Fig.~\ref{fig:dan}--\ref{fig:pada} (with class information) and Fig.~\ref{fig:dan_st}--\ref{fig:pada_st} (with domain information). We randomly select five classes in the source domain not shared with target domain and five classes shared with target domain. \textbf{(1)} Fig.~\ref{fig:dan}--\ref{fig:grl} show that the bottleneck features are mixed all together, meaning that DAN and DANN cannot discriminate both source and target classes very well; Fig.~\ref{fig:dan_st}--~\ref{fig:grl_st} show that the target data are aligned to all source classes including those outlier ones, which triggers negative transfer. \textbf{(2)} Fig.~\ref{fig:grl}--\ref{fig:rtn} show that RTN discriminates the source domain well but the features of most target data are very close to the source data even to the wrong source classes; Fig.~\ref{fig:rtn_st} further indicates that RTN tends to draw target data close to all source classes even to those not existing in target domain. Thus, their performance on target data degenerates due to negative transfer. 
\textbf{(3)} Fig.~\ref{fig:pada} and \ref{fig:pada_st} show that PADA can discriminate different classes in both source and target while the target data are close to the right source classes, while the outlier source classes cannot influence the target classes. These in-depth results show the versatility of the weighting mechanism. 

\section{Conclusion}
This paper presented a novel approach to partial domain adaptation. Unlike previous adversarial domain adaptation methods that match the whole source and target domains based on the shared label space assumption, the proposed approach simultaneously circumvents negative transfer by down-weighing the outlier source classes and promotes positive transfer by maximally matching the data distributions in the shared label space. Our approach successfully tackles partial domain adaptation problem where source label space subsumes target label space, testified by extensive experiments on various benchmark datasets.

\section{Acknowledgements}
This work is supported by National Key R\&D Program of China (2016YFB1000701), and National Natural Science Foundation of China (61772299, 61672313, 71690231).

\clearpage

\bibliographystyle{splncs}
\bibliography{1814}

\end{document}